\pdfoutput=1

\documentclass[conference]{IEEEtran}
\IEEEoverridecommandlockouts
\usepackage{cite}
\usepackage{amsmath,amssymb,amsfonts}
\usepackage{algorithmic}
\usepackage{graphicx}

\usepackage{float} 
\usepackage{multirow}

\usepackage{capt-of}

\usepackage{textcomp}
\usepackage{xcolor}
\usepackage[colorlinks,linkcolor=blue,anchorcolor=red,citecolor=green]{hyperref}
\usepackage[numbers,sort&compress]{natbib}

\def\BibTeX{{\rm B\kern-.05em{\sc i\kern-.025em b}\kern-.08em
    T\kern-.1667em\lower.7ex\hbox{E}\kern-.125emX}}
\begin{document}


\title{Weakly Supervised Two-Stage Training Scheme for Deep Video Fight Detection Model
{\footnotesize}
  \thanks{*These authors contributed equally to this work.}
}

\makeatletter
\newcommand{\linebreakand}{
  \end{@IEEEauthorhalign}
  \hfill\mbox{}\par
  \mbox{}\hfill\begin{@IEEEauthorhalign}
}
\makeatother

\author{
\IEEEauthorblockN{Zhenting Qi*}
\IEEEauthorblockA{\textit{ZJU-UIUC Institute} \\
\textit{Zhejiang University}\\
Haining, China \\
Zhenting.19@intl.zju.edu.cn}
\and
\IEEEauthorblockN{Ruike Zhu*}
\IEEEauthorblockA{\textit{Department of Computer Science} \\
\textit{University of Illinois at Urbana-Champaign}\\
Urbana, IL, USA \\
ruikez2@illinois.edu}
\and
\IEEEauthorblockN{Zheyu Fu*}
\IEEEauthorblockA{\textit{ZJU-UIUC Institute} \\
\textit{Zhejiang University}\\
Haining, China \\
Zheyu.19@intl.zju.edu.cn}\\
\linebreakand 
\IEEEauthorblockN{Wenhao Chai*}
\IEEEauthorblockA{\textit{ZJU-UIUC Institute} \\
\textit{Zhejiang University}\\
Haining, China \\
Wenhaochai.19@intl.zju.edu.cn}
\and
\IEEEauthorblockN{Volodymyr Kindratenko}
\IEEEauthorblockA{\textit{National Center for Supercomputing Applications} \\
\textit{University of Illinois at Urbana-Champaign}\\
Urbana, IL, USA \\
0000-0002-9336-4756}
}

\maketitle

\begin{abstract} Fight detection in videos is an emerging deep learning application with today's prevalence of surveillance systems and streaming media. Previous work has largely relied on action recognition techniques to tackle this problem. In this paper, we propose a simple but effective method that solves the task from a new perspective: we design the fight detection model as a composition of an action-aware feature extractor and an anomaly score generator. Also, considering that collecting frame-level labels for videos is too laborious, we design a weakly supervised two-stage training scheme, where we utilize multiple-instance-learning loss calculated on video-level labels to train the score generator, and adopt the self-training technique to further improve its performance. Extensive experiments on a publicly available large-scale dataset, UBI-Fights, demonstrate the effectiveness of our method, and the performance on the dataset exceeds several previous state-of-the-art approaches. Furthermore, we collect a new dataset, VFD-2000, that specializes in video fight detection, with a larger scale and more scenarios than existing datasets. The implementation of our method and the proposed dataset will be publicly available at \href{https://github.com/Hepta-Col/VideoFightDetection}{https://github.com/Hepta-Col/VideoFightDetection}.

\end{abstract}


\section{Introduction}

\textbf{Video fight detection (VFD)} aims to detect and localize fight events in videos temporally or spatially. With the prevalence of surveillance cameras and streaming media, VFD is getting increasingly important as it serves as an automatic tool to monitor and identify aggressive scenes without human intervention, thus largely reducing the cost. For example, online videos can be automatically checked and tagged with age restrictions if containing violent scenes. Also, surveillance systems equipped with VFD modules can provide a real-time warning to the police and people around.

VFD has been researched for years due to its practical prospect. Early studies on VFD \cite{articleCNNLSTM, Li2018, estmm} largely rely on \textbf{Action Recognition} techniques, since human fights can be intuitively considered as a subset of human actions. These approaches usually use a video encoder to obtain action-aware features and then apply a classifier to the features to decide whether normal or not. Therefore, the encoders are usually trained on very short video clips \cite{cheng2021rwfdataset, reallifedataset} due to the device memory limitation on one batch of videos. Although these models show good performance in terms of various binary classification metrics, they ignore the temporal dependencies between time periods of an event happening in real life. 

Recently, with the development of \textbf{Video Anomaly Detection (VAD)} methods, some researchers tried treating VFD as a subtask of VAD \cite{degardin2021iterative, tan2022detection}. In this sense, VFD models are trained to learn the temporal context information of videos, therefore they are able to detect anomalies precisely even if given long videos. But the training process shares the same challenge with VAD that it is hard for models to fully learn the inherent discrepancies between normal videos and abnormal videos. Moreover, fight events differ from general video anomalies in that they usually happen in short periods, which poses an even greater challenge for the model to detect them.

Considering these aspects of VFD, our idea to construct a model and design a training scheme is two-fold: first, we take into account action recognition and anomaly detection together; second, we train the model on weak labels, so that collecting training data would not require too much human labor. To this end, we develop the model as a composition of a feature extractor head and an anomaly score generator tail; next, we utilize a two-stage training framework, where multiple-instance-learning loss is applied to train the score generator from scratch and pseudo labels are generated to supervise another training iteration of it. Experiments show that our model outperforms the current video anomaly detection state-of-the-art (SOTA) models on the UBI-Fights dataset by reaching AUROC of 93.1\%, which is 1.2 points higher than current SOTA models reported in the literature.

Existing fight detection datasets are limited in size and scenarios. To address this issue, we construct VFD-2000, a large-scale, multi-scenario dataset with both video-level and frame-level labels. Our dataset has the following advantages over others: i) containing fight behaviors in multiple scenarios, reducing the effect of background in fight detection; ii) containing the corresponding non-fight videos under the same scene as fight videos, eliminating the background differences between fight and non-fight videos; iii) including clips that are showing ambiguous behaviors between fight and non-fight; iv) containing both short and long, vertical- and horizontal-view videos for video-level weakly supervised and supervised training.

The main contributions of our work are:
\begin{itemize}
  \item We propose an effective deep learning model dedicated for video fight detection.
  \item We propose a two-stage training framework to train the fight detection model from coarse to fine in a weakly-supervised manner.
  \item Our experimental results on the UBI-Fights dataset outperform current SOTA models.
  \item We collect a large-scale, multi-scenario dataset with both video-level and frame-level labels made specifically for video fight detection tasks.
  \item We conduct several benchmark tests on our proposed VFD-2000 and report SOTA results.
\end{itemize}

\section{Related Work}
\subsection{Fight Detection}

Fight detection in videos is usually considered a binary classification task. A large body of work is based on a video feature extractor followed by a classifier that outputs \textit{fight} or \textit{no fight} label for each video. The features of interest typically contain spatial-temporal and/or motion information. Zhang et al. \cite{Zhang2015} use the Gaussian Model of Optical Flow to extract candidate violence region, and then the Orientation Histogram of Optical Flow is applied to classify the violent behavior. Ditsanthia et al. \cite{CCTV} propose an approach that utilizes the convolutional neural network (CNN) as a feature extractor, and also present a new kind of visual feature descriptors to describe visual information in a video frame. Ullah et al. \cite{Ullah2019} implement a CNN-based model to recognize objects and remove unwanted frames. Other works also utilize action recognition techniques to solve fight detection tasks. Accatoli et al. \cite{articleCNNLSTM} apply a pre-trained 3D CNN to extract motion features. The authors use these descriptors as an input for a linear SVM to classify videos as violent or non-violent. Li et al.\cite{Li2018} combine both the RGB-based model and the optical flow Graph Model, and feed these two kinds of images into a 3D-CNN model to get the final results. Kang et al. \cite{estmm} propose a pipeline that combines spatial attention with the 2D-CNN model to extract features and use a Temporal Squeeze-and-Excitation block to highlight the periods that might have fight events. Recently, Choqueluque-Roman et al. \cite{choqueluque2022weakly} utilize a Fast-RCNN-based architecture to realize violence detection in a weakly supervised manner, with modules designed especially for human actions and dynamics. However, we notice that few researchers except Tan et al. \cite{tan2022detection} and Degardin et al. \cite{degardin2021iterative} treat fight detection as an anomaly detection task. 

\subsection{General Video Anomaly Detection}
Given a video, a VAD model is supposed to localize the anomaly time span(s). Here anomaly is defined as a deviation from what is typically expected based on the majority of the video frames. However, obtaining frame-level annotations is usually too laborious. Hence, most studies on VAD focus on unsupervised or weakly supervised methods. Unsupervised methods \cite{hasan2016learning, zhou2019anomalynet, rodrigues2020multi, cai2021appearance} usually train models using purely normal data, and the model is expected to recognize the distinctive data as abnormal data. For weakly supervised methods \cite{sultani2018real, zhong2019graph, lv2021localizing, tian2021weakly, feng2021mist}, the video-level labels (normal/abnormal) are available. In this scenario, the task is dubbed as {\bf{Weakly Supervised Video Anomaly Detection (WSVAD)}}. 

Past work on WSVAD has tried several ways to leverage these weak labels. Sultani et al. \cite{sultani2018real} first propose {\bf{multiple-instance-learning (MIL)}}, where each video is treated as a video segment bag and the model is trained to maximize the separability between the normal bags and abnormal bags. However, this method suffers from noisy scores and the inability to consider the temporal relation between clips. Motivated by these drawbacks, later studies concentrate on improving the MIL-training scheme. Zhong et al. \cite{zhong2019graph} treat WSVAD as a supervised learning task with noisy labels, and use a GCN-based model to clear the noisy labels; Lv et al. \cite{lv2021localizing} propose a high-order context encoding model to utilize the temporal context, and apply a new strategy to facilitate noise inference; RTFM model proposed by Tian et al. \cite{tian2021weakly} uses a multi-scale temporal network (MTN) to encode both long-range and short-range temporal relations between video clips, greatly improving the robustness of MIL-based model training. And in MIST \cite{feng2021mist}, a modified MIL loss is used to train a score generator, which then outputs clip-level pseudo labels to supervise the training of a self-guided attention module. 

\subsection{Self-training}
Self-training has been investigated extensively in semi-supervised learning. Much of the past work on self-training focused on {\bf{pseudo-labeling strategies}}. At each iteration, self-training selects a portion of the unlabeled training data for pseudo-labeling. An early approach sets the selecting threshold to the average of respectively positive and negative predictions \cite{tur2005combining}. Lee et al. \cite{lee2013pseudo} leverage neural networks as a supervised classifier and pick up the class with the maximum predicted probability to infer pseudo-labels. In more recent work, Iscen et al. \cite{iscen2019label} implement pseudo-labeling through graph-based label propagation. They make predictions on the entire dataset by employing a transductive label propagation method and use the predictions to generate pseudo-labels.

Besides leveraging single classifiers in self-training algorithms, some studies have proposed to utilize two classifiers, where each model learns on the output of the other. This method is also called co-teaching, in which two classifiers successively switch roles to assign pseudo-labels for the other to train. Xie et al. \cite{xie2020self} show the simplicity and effectiveness of co-teaching to leverage unlabeled data. They propose a model with manually injected noises in the student model to learn beyond the teacher’s knowledge. Chen et al. \cite{chen2021semi} propose to use cross pseudo supervision to regularize and enforce the consistency between the two segmentation networks. 

\section{Methods}

\subsection{Problem Formulation} 
We consider Video Fight Detection as an intersection of Action Recognition and Video Anomaly Detection. Different from most of the previous works that treat VFD as a classification task, we formulate it as a regression task that relies on scores to define the output. Plus, since the annotations are video-level binary labels, i.e. \textit{normal} (does not contain fight event) and \textit{abnormal} (contains at least one fight event), our problem formulation largely follows the existing WSVAD work \cite{sultani2018real,zhong2019graph,lv2021localizing,tian2021weakly,feng2021mist}: given a video $V=\{v_i\}_{i=1}^N$ consisting of N clips, each of which has $T$ frames, along with a video-level label $Y \in \{0, 1\}$ indicating whether the video contains any fight event or not, the model is trained and is expected to calculate anomaly scores $S=\{s_i\}_{i=1}^{N'}$ for an unseen video, with each clip $v_i$ given a score $s_i \in [0, 1]$ indicating the probability of a fight event occurring. 

\subsection{Video Fight Detection Model} 
To combine action recognition and anomaly detection techniques, we propose a fight detection model composed of a feature extractor head $f$ and an anomaly score generator tail $g$, as shown in Fig.~\ref{fig:detection_model}. The feature extractor encodes the video with special attention to human actions: accepting a video $V=\{v_i\}_{i=1}^N$ broken up into $N$ clips, it outputs features $X=\{x_i\}_{i=1}^N$ corresponding to all clips, where:
\begin{equation}
    x_i=f(v_i),
\end{equation}
then the score generator is supposed to rate the clips in terms of their relevance to fight: it takes the extracted features as input and calculates the anomaly scores $S=\{s_i\}_{i=1}^N$ for each clip of the video:
\begin{equation}
    S=g(X),
\end{equation}
where the score of every clip is calculated according to the features $X$ of the whole group of clips, whereby temporal relation between clips is injected into each score.

\begin{figure}[]
    \centering
    \includegraphics[width=9.3cm]{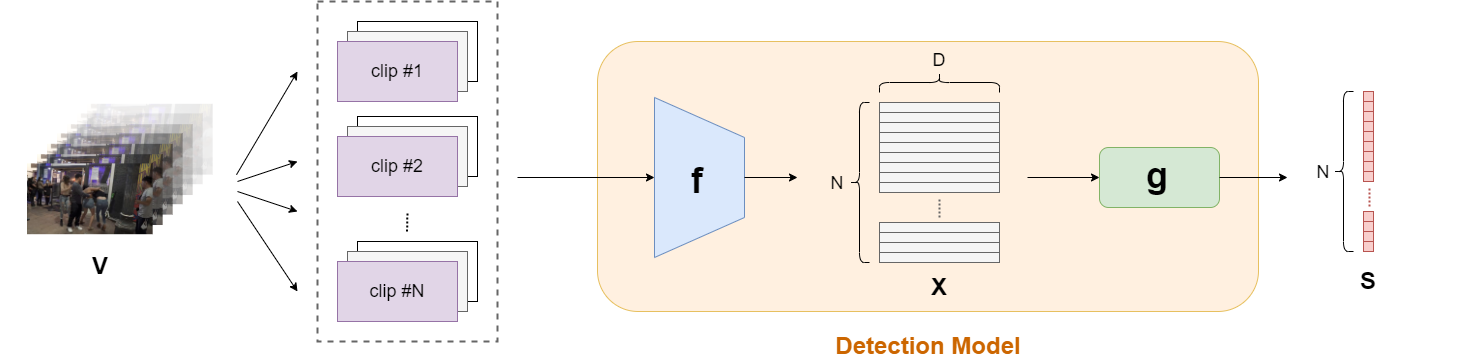}
    \caption{Workflow \& structure of the fight detection model.}
    \label{fig:detection_model}
\end{figure}

\subsection{Two-stage Training Framework}
Next, to train such a model on weak labels, we design a two-stage training framework as shown in Fig.~\ref{fig:two_stage}. To distinguish between these two stages, we name the first generator as network A and the second as network B, while keeping the feature extractor unchanged. Theoretically, this gives rise to two scenarios: A equals B, and A differs from B. We compare these two scenarios in Section IV. 

\begin{figure}[]
    \centering
    \includegraphics[width=9cm]{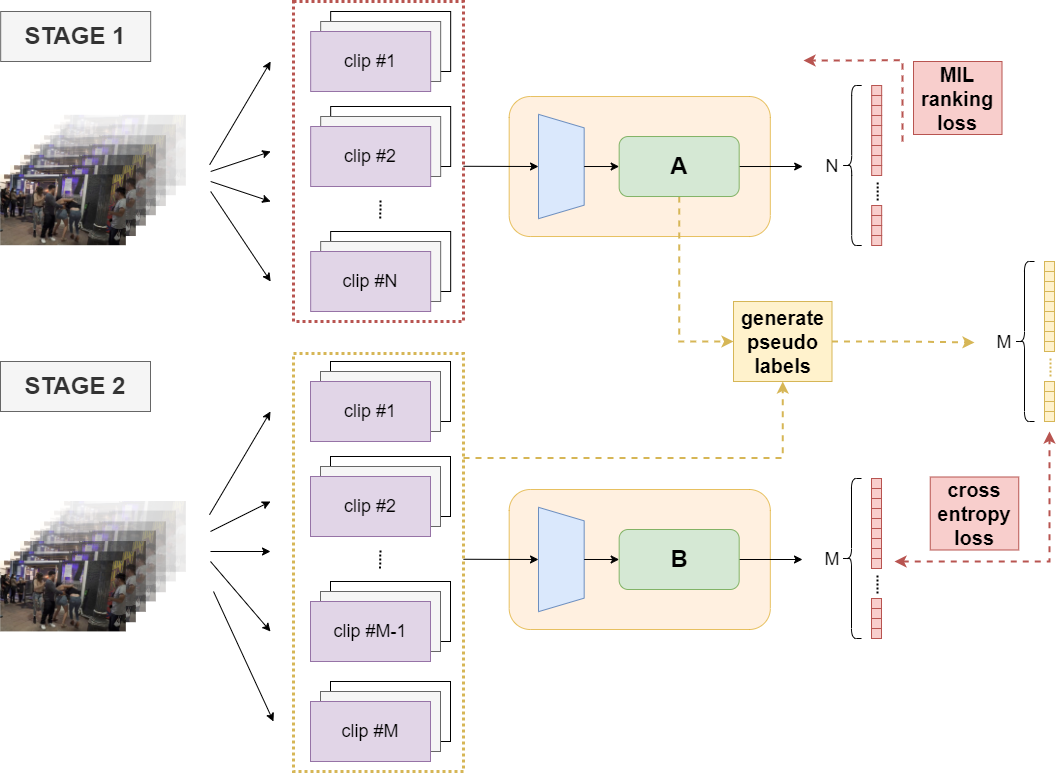}
    \caption{Two-stage training framework. At stage 1, each video is cut into N clips (N is a constant), and Network A is trained with MIL loss (note that if A $\neq$ B, B should also be trained from scratch at stage one); after stage 1 is done, each video is cut into M clips so that each has exactly $T=32$ frames (M is a variable depending on video length), and pseudo labels for each clip are generated; at stage 2, the pseudo labels are used to supervise Network B's training.}
    \label{fig:two_stage}
\end{figure}

\subsubsection{Stage One: Score Generator Training via Multiple Instance Learning} 
To train the anomaly score generator A from scratch, we follow previous WSVAD work \cite{sultani2018real,zhong2019graph,lv2021localizing,tian2021weakly,feng2021mist} to use an MIL ranking loss as constraint that maximizes the separability between normal and abnormal videos. Given a pair of normal video $V^n=\{v^n_i\}_{i=1}^N$ and abnormal video $V^a=\{v^a_i\}_{i=1}^N$ (``a" means ``abnormal" and ``n" means ``normal"), the feature extractor $f$ first encodes the clips into $D$-dimensional feature vectors $X^n=\{x_i^n\}_{i=1}^N$ and $X^a=\{x_i^a\}_{i=1}^N$: 
\begin{equation}
    x_i^n=f(v_i^n),\ x_i^a=f(v_i^a),
\end{equation}
where $x_i \in \mathbb{R}^D$. Then generator A calculates scores $S^n$ and $S^a$ from the features:
\begin{equation}
    S^n=g_A(X^n),\ S^a=g_A(X^a),
\end{equation}
where $S=\{s_i\}_{i=1}^N$ are scalar values between 0 and 1. Using these scores, an MIL loss function can be defined as:
\begin{equation}
    \mathcal{L}_{MIL} = \max(0, \epsilon-d(S^n, S^a)),
\end{equation}
where $\epsilon$ is a fixed margin, and function $d$ measures the distance, or separability, between normal and abnormal videos. The instantiation of $d$ depends on the choice of the generator. Following existing WSVAD work \cite{sultani2018real,zhong2019graph,lv2021localizing,tian2021weakly,feng2021mist}, we add the sparsity and smoothness constraints to the loss. Note that the feature extractor is not involved in the parameter updating procedure; and if A $\neq$ B, B should also be trained from scratch at stage one, following the same manner as training A.

\subsubsection{Stage Two: Score Generator Refining with Pseudo Labels} 
Inspired by previous self-training methods in the field of semi-supervised learning \cite{feng2021mist, xie2020self, chen2021semi}, we apply a self-training scheme at stage two. After the first stage is done, we break every video in the training set into M equal-length clips $V=\{\hat{v_i}\}_{i=1}^M$ (here M is a variable depending on every video length) and pass them through the model again:
\begin{equation}
    \hat{x_i}^n=f(\hat{v_i}^n),\ \hat{x_i}^a=f(\hat{v_i}^a),
\end{equation}
\begin{equation}
    \hat{S}^n=g_A(\hat{X}^n),\ \hat{S}^a=g_A(\hat{X}^a),
\end{equation}
\begin{equation}
    \hat{Y}=h(\hat{S}),
\end{equation}
where $\hat{X}= \{\hat{x_i}\}_{i=1}^M$, $h$ is a transformation of raw scores $\hat{S}=\{\hat{s_i}\}_{i=1}^M$, and $\hat{Y}=\{\hat{y}_i\}_{i=1}^M (\hat{y_i}\in[0, 1])$ are the generated pseudo labels. Note that we use soft pseudo labels here to prevent over-confidence, and pseudo labels are generated only for abnormal videos since we can safely assign 0 to all clips of any normal video. Then following the same workflow as A, the generator B generates another set of scores $S'$ by:
\begin{equation}
    S'^n=g_B(\hat{X}^n),\ S'^a=g_B(\hat{X}^a),
\end{equation}
and a cross-entropy loss is applied to the pseudo labels $\hat{Y}$ and scores $S'$. After two stages are done, only generator B is used for inference.

The motivation for applying a two-stage training scheme here is two-fold: First, the transition from video-level learning to clip-level learning is a coarse-to-fine learning procedure, which helps the model focus on increasingly shorter time spans. At stage one any video would be broken up into a fixed number of clips regardless of its original length ($N$ clips for each video), therefore the scores assigned are coarse-grained, i.e. a long span of frames share a single score; at stage two, however, the videos are cut up in a way that each clip has identical length $T$ ($T$ is far less than each video length), so more clips will be generated while each clip has a smaller number of frames, which means the scores generated are fine-grained, i.e. scores are assigned to shorter spans of frames. Second, the model gets stronger by learning from faithful predictions generated by itself. If the model is trained powerful enough on the video-level supervision, we can predict that through learning on the high-quality clip-level pseudo labels generated by itself, the model could perform even better in generating anomaly scores. 

\subsection{Dataset Collection}


\subsubsection{Existing Fight Detection Datasets}

So far, several datasets have been used in fight detection. The videos in those datasets are given video-level or frame-level labels (`Fight' and `NonFight'). Those datasets have several drawbacks such as fixed video length, small scale, limited scenes, etc. For example, the small-scale problem appears in the movie dataset \cite{hockydataset} and surveillance camera fight dataset \cite{surveillancedataset}. They contain 200 videos and 300 videos, respectively, and each video in these two datasets lasts for about 1 to 2 seconds. The hockey dataset \cite{hockydataset} is limited in the variety of scenes. It's sampled based on ice hockey games videos and does not include fight behaviors in other real-life situations. The CCTV-Fight dataset \cite{cctvdataset} only includes fight videos and does not provide non-fight samples. The UCF-Crime dataset \cite{sultani2018real} is large-scale with 1900 long videos where each video is longer than 60 seconds, but it's not specifically focused on fight detection. The RLV \cite{reallifedataset} and RWF \cite{cheng2021rwfdataset} dataset both contain 2000 videos which are large in number, but they only contain short video clips where the length of videos is in the range of 3 to 7 seconds. The UBI-Fights dataset \cite{ubidataset} contains 1000 videos with a total of 80 hours and is fully annotated at the frame level. It includes videos from different recording angles such as moving, rotating, and stationary cameras. We will use it as a comparison dataset in our experiments.

\subsubsection{Data Collection}

We propose \textbf{VFD-2000}, a video fight detection dataset containing more than 2000 videos. We use YouTube as the data source. We search for specific scenarios and use ``fight" as a search keyword, for example, ``street fight", ``beach fight", and ``violence in the restaurant". We collect 200 videos under 20 different scenes. Some of those videos are collections of very short fight videos, and most of the videos are recorded by smartphone or moving camera device, therefore the scenes in those videos vary significantly. In order to keep the scene in the video to be more consistent, we use the PySceneDetect package~\cite{scenedetect} to cut the video into the scene clips. We set the threshold of the scene detector to be 30 and min clip length to 2 seconds. After the clipping, we get about 3500 video clips with a time length from 2 seconds to 3 minutes. But there are still many unnecessary clips containing video introductions, transition animations, or black screen, having shaky video angles or blurred picture quality, or some special filters added by the video producer. To eliminate those clips we manually examine the videos and add the video-level labels to each clip. After the filtering, there are 2490 video clips left: 1296 of them are fight videos, and 1194 are non-fight videos. We then divide our dataset into four categories: long video and short video in vertical and horizontal views. Those longer than 96 frames will be counted as long videos, while others will be viewed as short videos. The horizontal view refers to those having their width longer than the height. The ratio of vertical and horizontal videos in our dataset is about 1:2, and we also keep this ratio when splitting the training and testing data. Fig.~\ref{fig:youtube_dataset_sample} shows a few  example frames for different scenes. Table \ref{tab:four_categories} shows the four categories and corresponding number of videos. We also provide the distribution of length of videos for our dataset in Fig.~\ref{fig:dist_of_data}, and compare our dataset with previous datasets in Table \ref{tab:datasets}.

\begin{table}[]
\caption{Four categories of video data in VFD-2000}
\begin{center}
\begin{tabular}{c|c|c}
\hline
 & \textbf{long video} & \textbf{short video} \\
\hline
vertical view&495&251\\
\hline
horizontal view&1057&687\\
\hline
\end{tabular}
\label{tab:four_categories}
\end{center}
\end{table}

\subsubsection{Data annotation}
For the annotation, we add video-level labels for each video manually. For long videos, We manually add frame-level labels to those fight videos in test data to test our model. For short videos, we use the video-level label in training and testing. We save the predicted results into a JSON file in a dictionary format with each video containing the predicted results by our model, scene labels as well as the URL from which this video was downloaded.

Compared with previous datasets, our dataset covers more real-life scenes. It contains fight behaviors in 20 different scenarios with few or multiple people. Among those scenes, we try to find both moving and stationery background videos in each scene. Besides, our dataset is sampling based on the whole video therefore most non-fight videos will be the corresponding period before and after the fight. The scenes of these non-fight videos will be the same as the corresponding fight videos, which can reduce the effect of background differences between fight and non-fight behaviors. Due to this sampling method, our dataset also includes clips that are showing ambiguous behaviors between fight and non-fight, such as throwing punches but not hitting each other, running at each other but not touching each other, or being pulled away after the fight is over. We also separate our dataset into four categories based on video length and size for supervised and weakly-supervised training.

\begin{table}[htbp]
\caption{Comparison with previous fight detection related datasets}
\scriptsize
\begin{center}
\begin{tabular}{p{1.9cm}|p{0.7cm}p{0.8cm}p{0.8cm}p{1.2cm}p{1cm}}
\hline
\textbf{Name} & \textbf{\#Videos} &\textbf{Length(sec)} &\textbf{Resolution}& \textbf{Annotation}&\textbf{Scenario}\\
\hline
CCTV-Fights \cite{cctvdataset}&1000& 5\~{}720 & variable & frame-level & CCTV\\

Hockey Fight \cite{hockydataset}& 1000 & 1.6\~{}1.96 & 360x288 & video-level & 	Hockey\\
Movies Fight \cite{hockydataset}& 200 & 1.6\~{}2 & 720x480 & video-level & movies\\
UCF-Crime \cite{sultani2018real}& 1900 & 60\~{}600 & variable & video-level & 	surveillance\\
UBI-Fights \cite{ubidataset}& 1000 & 1\~{}600 & 640x360 & both & real life \\

RLV \cite{reallifedataset}& 2000 & 3\~{}7 & variable & video-level & real life\\   
RWF \cite{cheng2021rwfdataset}& 2000 & 5 & variable & video-level & surveillance\\   
Surveillance \cite{surveillancedataset}& 300 & 2 & variable & video-level & surveillance\\
\hline


\multirow{2}*{Ours$^{a}$} & 938(short) & 1\~{}3.2 & variable& video-level & \textbf{multiple$^{b}$} \\
		\cline{2-6}
		~ & 1552(long) & 3.2\~{}208 & variable& both & \textbf{multiple$^{b}$}\\

\hline
\multicolumn{6}{l}{$^{\mathrm{a}}$ We split our dataset into long and short videos by number of video frames.} \\
\multicolumn{6}{l}{$^{\mathrm{b}}$ Our dataset includes 20 different scenes, Fig.~\ref{fig:dist_of_data} gives detailed description.} \\
\end{tabular}
\label{tab:datasets}
\end{center}
\end{table}



\begin{figure*}[]
    \includegraphics[width=1\linewidth]{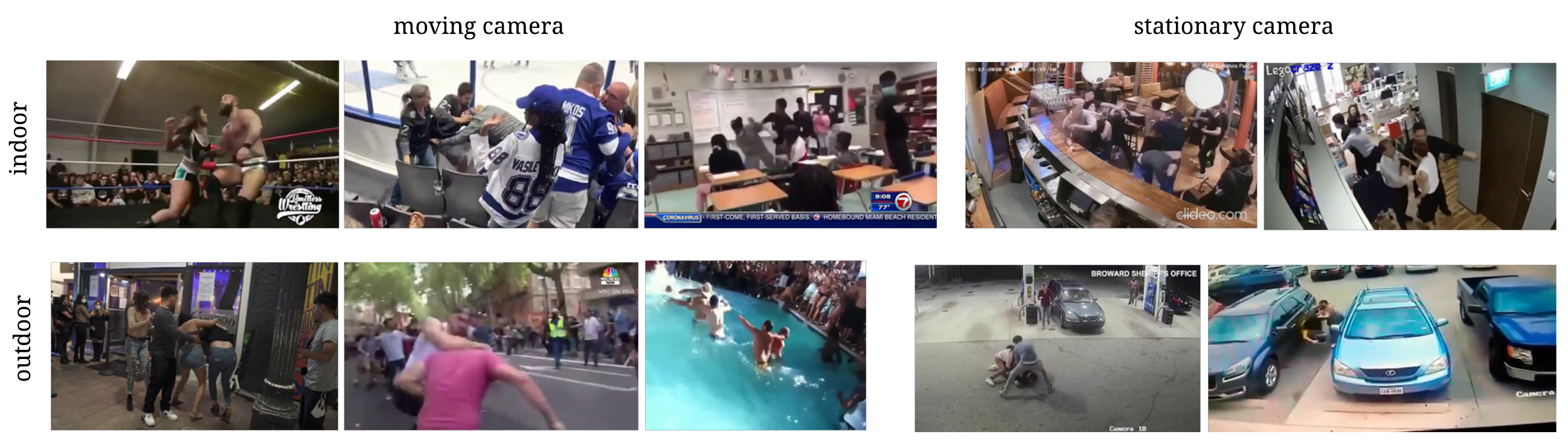}
    \caption{VFD-2000 consists of outdoor and indoor videos recorded by both stationary and moving cameras.}
    \label{fig:youtube_dataset_sample}
\end{figure*}

\begin{figure}[]
    \includegraphics[width=9.6cm]{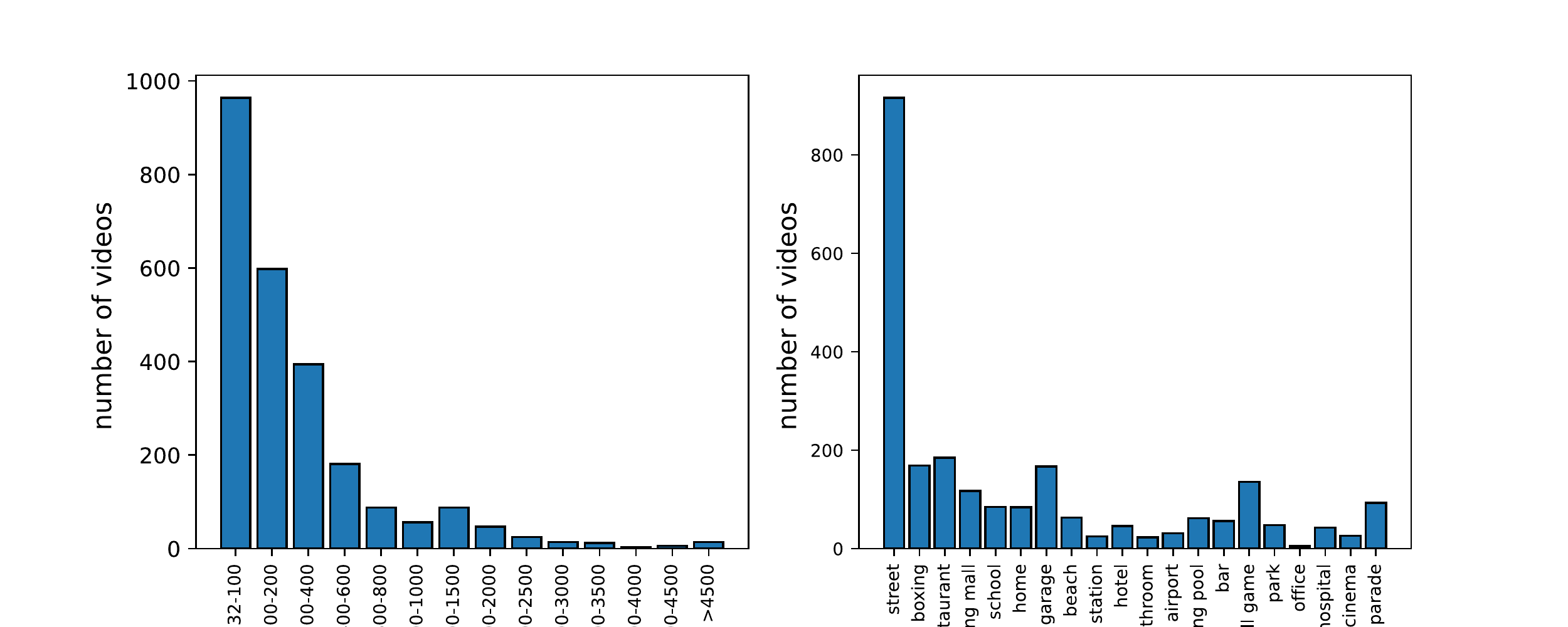}
    \caption{The left bar chart shows the distribution of our dataset by number of frames. The right bar chart shows the number of videos for each scene, among all these scenario street fight account for the majority of videos.}
    \label{fig:dist_of_data}
\end{figure}

\section{Results and Discussion}
\subsection{Implementation Details}
We choose I3D model \cite{carreira2017quo} pretrained on Kinetic400 \cite{kay2017kinetics} as the action-aware feature extractor. Before detection model training, we first pre-process the videos by converting raw mp4 format to npy format and resizing them to 224$\times$224 pixels. Then we cut the videos into clips with $T=32$ frames per clip, which are then passed through the I3D model one by one. The clip-level features are extracted from ``mix\_5c" layer of I3D ($D=2048$), grouped by video name, and stored as npy files.

In the first stage, we follow Sultani et al. \cite{sultani2018real} to cut each video into $N=32$ clips, regardless of its original length. To realize this, for each video we average every $\frac{1}{N}$ portion of clip features of the whole span to get new $N$ clip features. Then these $N$ new features of each video are passed to the detection model as a whole. In the second stage, clip features of each video are directly passed through the detection model without ensuring that the number of clip features is constant (since videos have different lengths).

As mentioned in Section III, we have tried different implementations of generator A and B (details are shown in Part D). We experiment with \textbf{Multi-scale Temporal Network (MTN)}, a dilated convolution-based module proposed by RTFM \cite{tian2021weakly} (current SOTA model on WSVAD task), and MLP, a simple regressor consisting of 3 layers with the number of units being 512, 32, and 1. During first-stage training, both MTN and MLP are trained for 10000 epochs with Adam optimizer (learning rate $=0.001$). Then for self-training, MTN is trained for 50 epochs with Adam optimizer (learning rate $=0.00001$) while MLP is trained for 200 epochs with Adam optimizer (learning rate $=0.001$). The score transformation, $h$, is instantiated by identical mapping and min-max normalization.

The entire pipeline is implemented in PyTorch, and the calculations are carried out on HAL system \cite{HAL}.

\subsection{Datasets and Metrics}
We conduct experiments mainly on UBI-Fights and our VFD-2000. UBI-Fights is a large-scale video anomaly detection dataset especially focusing on fights, which contains 80 hours of video fully annotated at the frame levels. Of its total 1000 videos, 784 of them are shot from normal daily life while 216 of them contain a fight event. All videos are resized to be 640x360 pixels, with the frame rate set to 30 fps. Importantly, unnecessary video segments, such as video introductions and text overlays that could disturb the learning process are removed. 

Following previous work, we test our model in terms of \textbf{Area Under the Receiver Operating Characteristic (AUROC)} and \textbf{Equal Error Rate (EER)}. AUROC is a widely used performance metric in classification tasks, which tells about the model’s ability to discriminate between positive examples and negative examples. It is calculated as the area under the ROC, showing the trade-off between true positive rate and false positive rate across different decision thresholds. EER is the location on a ROC curve where the false positive rate and false negative rate are equal. Generally speaking, the lower the EER is, the higher accuracy the model can achieve. 

\subsection{Comparison with SOTA Models}

We compare our model's performance with previous SOTA video anomaly detection and human action recognition models on UBI-Fights and our VFD-2000. For fairness, we present experimental results for both AUROC and EER. 

\subsubsection{On UBI-Fights Dataset}
Results of UBI-Fights are shown in Table \ref{tab:sota}, showing that our method surpasses all the previous approaches. For AUROC, our method reaches \textbf{93.1\%}, which is at least \textbf{1.2 points} higher than current SOTA models. It is worth mentioning that the previous best performance (AUROC=91.9\%) produced by Tan et al. \cite{tan2022detection} uses 10-crop video preprocessing for model training, which is very time- and space-consuming, while our model is trained with just 1-crop. Meanwhile, the EER of our model predictions fells below 10\% for the first time. 

\subsubsection{On the Collected VFD-2000 Dataset}
We also conduct experiments on the proposed VFD-2000, as shown in Table \ref{tab:sota_vfd}. We compare our model with previous video anomaly detection, human action recognition, and violence detection models trained by us on our dataset. On VFD-2000-short, we train C3D \cite{c3d} and TimeSformer \cite{timesformer} with video-level labels based on MMAction2 \cite{mmaction2} framework. We also trained the ESTMM \cite{estmm} pipeline. Among those models, the TimeSFormer model reaches the highest accuracy at 79.79\%. Note that since WSVAD methods are designed for detecting anomalies in long videos, we did not run our model or any WSVAD-based model on VFD-2000-short. On VFD-2000-long, where we compare WSVAD methods, our model exceeds RTFM, the current SOTA model on WSVAD, by 1.49 points in terms of AUROC, and decreases EER to 40.55\%.

\begin{table}[]
\caption{Experimental results on UBI-Fights}
\begin{center}
\begin{tabular}{cc|cc}
\hline
\textbf{Method} & \textbf{Supervised} & \textbf{\textit{AUROC (\%)}} & \textbf{\textit{EER (\%)}} \\
\hline
Hasan et al. \cite{hasan2016learning}&Un.&52.8&46.6\\

Adv. Generator \cite{ravanbakhsh2017abnormal}&Un.&53.3&48.4\\

LSTM-AE \cite{chong2017abnormal}&Weak&54.1&48.0\\

s2-VAE \cite{wang2018generative}&Un.&61.0&42.7\\

Degardin et al. \cite{degardin2021iterative}&Weak&84.6&21.6\\

Sultani et al. \cite{sultani2018real}&Weak&89.2&18.6\\

RTFM$^{a}$ \cite{tian2021weakly}&Weak&90.4&12.6\\

GMM \cite{degardin2020weakly}&Weak&90.6&16.0\\

Tan et al.$^{b}$ \cite{tan2022detection}&Weak&91.9&-\\
\hline
\textbf{Our model}$^{a}$&Weak&\textbf{93.1}&\textbf{8.5}\\
\hline
\multicolumn{4}{l}{$^{\mathrm{a}}$Using 1-crop. \quad $^{\mathrm{b}}$Using 10-crop.}\\
\end{tabular}
\label{tab:sota}
\end{center}
\end{table}

\begin{table}
	\centering
	\caption{Experimental results on VFD-2000}
	\begin{tabular}{c|c|ccc}
        \hline
        \textbf{Split} & \textbf{Method} & \textbf{\textit{AUROC (\%)}} & \textbf{\textit{EER (\%)}} & \textbf{\textit{Acc (\%)}} \\
		\hline
		\multirow{3}*{short$^{a}$} & C3D~\cite{c3d} & \textbf{83.88} & 26.56 & 73.40\\
		~ & TimeSformer~\cite{timesformer} & 82.19 & \textbf{25.60} & \textbf{79.79}\\
		~ & ESTMM~\cite{estmm} &78.30 & 33.01 & 73.40\\
		\hline
		\hline
		\multirow{3}*{long$^{b}$} &RTFM \cite{tian2021weakly} & 61.20 & 41.44 & 69.01\\
		\cline{2-5}
		~ & Ours (MLP+MLP) & 53.00 & 46.29 & 22.57\\
		~ & Ours (MTN+MTN) & \textbf{62.69} & \textbf{40.55} & \textbf{72.75}\\
		\hline
		\multicolumn{5}{l}{$^{\mathrm{a}}$Models predict video-level labels for short videos} \\
		\multicolumn{5}{l}{$^{\mathrm{b}}$Models predict frame-level labels for long videos}\\
	\end{tabular}
	\label{tab:sota_vfd}
\end{table}

\subsection{Ablation Study}
Here we perform ablation studies to reveal the influences on the result brought by different model implementations. All the experiments are conducted on UBI-Fights.

\subsubsection{Two-stage Framework}

To verify the effectiveness of the two-stage framework, we implement both one-stage and two-stage schemes to see whether two-stage training improves performance. As shown in Table \ref{tab:two_stage}, compared with the one-stage framework, both MLP and MTN achieve higher AUROC scores under two-stage training. Specifically, MLP's performance increases by 1.30 points, and that of MTN increases by 2.71 points. 

\begin{table}[]
\caption{Comparisons between one-stage and two-stage frameworks on UBI-Fights}
\begin{center}
\begin{tabular}{c|c|cc}
\hline
\textbf{\# Stages} & \textbf{Models} & \textbf{\textit{AUROC (\%)}} & $\Delta (\%)$\\
\hline
One-stage (baseline)&MLP&80.51&-\\
Two-stage&MLP+MLP&81.81&+1.30\\
\hline
One-stage (baseline)&MTN&90.36&-\\
Two-stage&MTN+MTN&\textbf{93.07}&+2.71\\
\hline
\end{tabular}
\label{tab:two_stage}
\end{center}
\end{table}

\subsubsection{Score Generator Choice}
Here we compare different combinations of score generators, as Table \ref{tab:generator} shows. Note that all of Network B have been trained from scratch for one round before stage-two starts. Conclusions that can be drawn from the results include i) MTN generates better pseudo labels because when keeping Network B unchanged, Network A being MTN achieves higher AUROC than that being MLP; ii) MTN is a stronger score generator than MLP since when Network A is identical to Network B, MTN exceeds MLP for more than 10 points. iii) It is not suitable that a strong score generator learns on pseudo labels generated by a weak score generator. When Network A is MLP and Network B is MTN, the result seriously deteriorates. 

\begin{table}[]
\caption{Comparisons between different generator choices of two-stage training on UBI-Fights}
\begin{center}
\begin{tabular}{cc|c}
\hline
\textbf{Network A} & \textbf{Network B} & \textbf{\textit{AUROC (\%)}}\\
\hline
MLP&MLP&81.81\\

MTN&MLP&84.86\\

MLP&MTN&75.80\\

MTN&MTN&\textbf{93.07}\\
\hline
\end{tabular}
\label{tab:generator}
\end{center}
\end{table}

\subsubsection{Number of Training Rounds}
We have also tried a different number of rounds of training the detection model. Table \ref{tab:round} shows that running the entire two-stage framework with MLP for another round could increase its AUROC by 1.2 points, but after more rounds it reaches a limit; and running with MTN for two rounds actually decreases AUROC by 0.75 points. In this sense, we consider the degradation of the performance to be caused by over-fitting of the training set.

\begin{table}[]
\caption{Comparisons between one-round and two-round self-training on UBI-Fights}
\begin{center}
\begin{tabular}{c|c|cc}
\hline
\textbf{\# Rounds} & \textbf{Models} & \textbf{\textit{AUROC (\%)}} & $\Delta (\%)$\\
\hline
1 (baseline)&MLP+MLP&81.81&-\\
2&MLP+MLP&83.01&+1.20\\
3&MLP+MLP&83.08&+1.27\\
4&MLP+MLP&83.24&+1.43\\
5&MLP+MLP&83.22&+1.41\\
\hline
1 (baseline)&MTN+MTN&\textbf{93.07}&-\\
2&MTN+MTN&92.32&-0.75\\
\hline
\end{tabular}
\label{tab:round}
\end{center}
\end{table}

\subsubsection{Score-to-label Transformation} 
We instantiate $h$ with identical mapping and min-max normalization, as shown in Table \ref{tab:transformation}. It shows that for the MLP$+$MLP models, applying min-max normalization improves the AUROC by 1.21 points compared to applying an identical mapping. However, for the MTN+MTN models, using identical mapping performs much better than min-max normalization by reaching AUROC at 93.07\%.

\begin{table}[]
\caption{Comparisons between two choices of score transformations on UBI-Fights}
\begin{center}
\begin{tabular}{c|c|c}
\hline
\textbf{$h$} & \textbf{Models} & \textbf{\textit{AUROC (\%)}}\\
\hline
identical mapping&MLP+MLP&81.81\\
min-max normalization&MLP+MLP&83.02\\
\hline
identical mapping&MTN+MTN&\textbf{93.07}\\
min-max normalization&MTN+MTN&81.17\\
\hline
\end{tabular}
\label{tab:transformation}
\end{center}
\end{table}

\subsection{Discussion}
\subsubsection{Efficiency of Two-stage Training}
According to MIST \cite{feng2021mist}, the key of their two-stage training scheme is to train a task-specific feature encoder. Therefore, in stage two they are applying end-to-end training, where the feature encoder is also involved in the forward pass and backpropagation. However, this is time- and space-consuming and might lead to over-fitting considering the large number of learnable parameters of video feature encoders. To circumvent these two problems, we utilize self-training only to train the score generator while keeping the feature encoder unchanged, and it turns out that the second stage takes very short time to finish while the high performance of the model is maintained. Specifically, in the second stage, training the MTN+MTN model for 50 epochs takes around 24 minutes, and training the MLP+MLP model for 200 epochs takes less than 1 hour.
\subsubsection{Quality of Generated Anomaly Scores}
\begin{figure*}[]
    \centering
    \includegraphics[width=18.5cm]{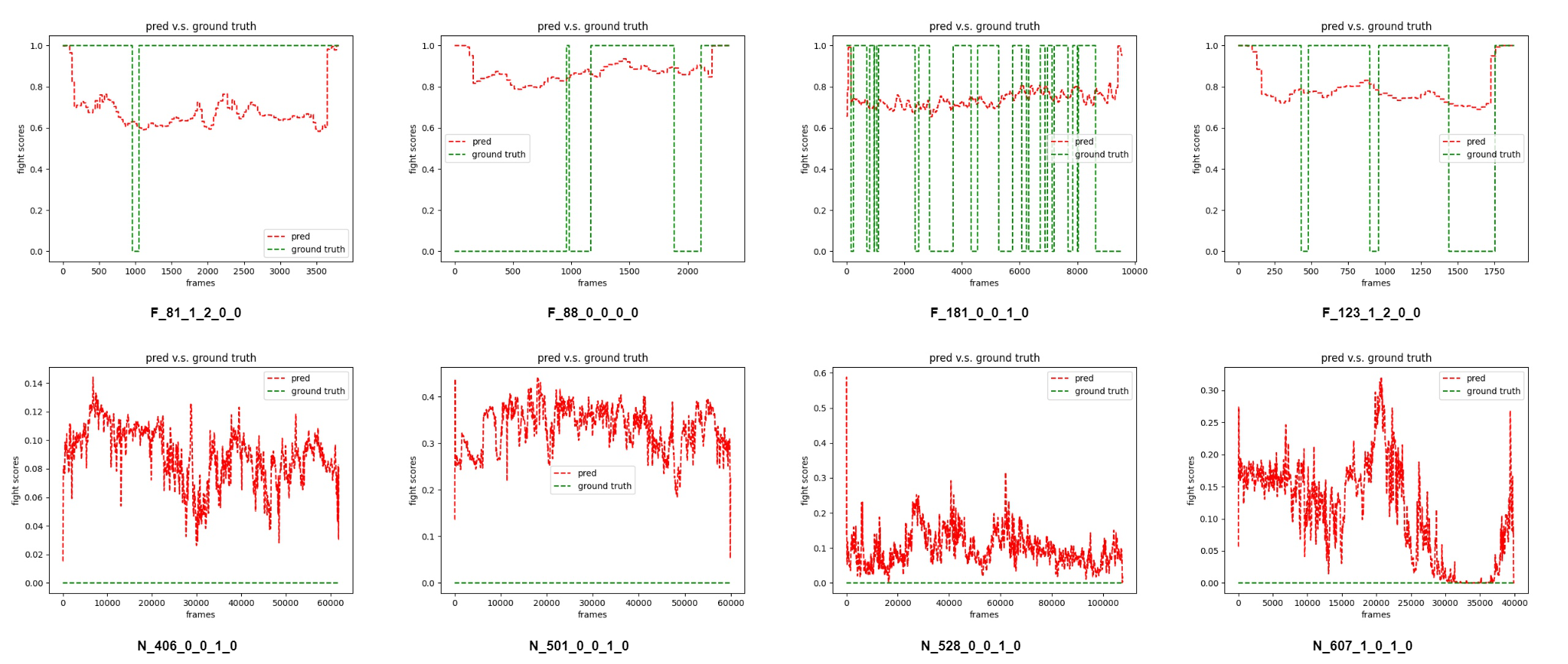}
    \caption{Examples of predictions vs. ground truths on UBI-Fights sample videos.}
    \label{fig:sci_figs}
\end{figure*}

We select four representative examples each for fight videos (upper row) and non-fight videos (lower row) in Fig.~\ref{fig:sci_figs}. The red curves stand for model predictions (anomaly scores), and the green curves are ground truth. The rising and dropping trends of the prediction curve fit well with the fluctuation of the ground truth curve. Also, it can be seen that the scores for fight videos are hovering at a high level, while that for non-fight videos are repressed at a low level. It might seem unsatisfactory to some extent that the absolute values of scores in fight videos lack discrimination between normal and abnormal clips, but this is a reasonable result since we use the smoothness loss on abnormal training samples, which exactly imposes restrictions on the fluctuation degree of the prediction curves, without which the prediction would become too noisy. 

\subsubsection{Generalization of Proposed Model}
Theoretically, if trained on different datasets, our model could be applied to any specific anomaly action detection task, such as detecting burglary, robbery, shooting, shoplifting, and arson. Plus, the feature extractor and the score generator could be replaced by other advanced models. For example, C3D \cite{c3d} could be a substitution for I3D.

\section{Conclusion}
In this work, we propose an effective model for video fight detection. To train such a model on weak labels and help it learn to focus on increasingly shorter time spans, we use a two-stage training framework to train the model from coarse to fine. Experimental results on the UBI-Fights dataset reveal that our model outperforms current SOTA models. Additionally, to better facilitate research on fight detection in videos, we collect a large-scale, multi-scenario dataset with both video-level and frame-level labels specifically for video fight detection tasks. Our experimental results, code, and dataset will be publicly available at https://github.com/Hepta-Col/VideoFightDetection.

\section*{Acknowledgment}

This work utilized resources supported by the National Science Foundation’s Major Research Instrumentation program, grant \#1725729, as well as the University of Illinois at Urbana-Champaign.

\bibliography{main.bib}


\end{document}